\documentclass[preprint,12pt,3p]{elsarticle}
\usepackage[T1]{fontenc}
\usepackage{lmodern}

\usepackage{amssymb}
\usepackage{amsmath}
\usepackage{graphicx}
\usepackage{booktabs}
\usepackage{siunitx}
\usepackage[hidelinks]{hyperref}
\usepackage{cleveref}

\myfooter[L]{Preprint\hfill\today}

\begin{document}

\begin{frontmatter}

\title{Fixing a Model That Learned Worse Cancer Means Lower Risk: Monotonic Constraints in Bladder Cancer Recurrence Prediction}

\author[newcastle]{Saram Abbas\corref{cor1}}
\ead{s.abbas11@newcastle.ac.uk}
\author[southampton]{David Thomas}
\author[urology]{Naeem Soomro}
\author[newcastle]{Rishad Shafik}
\author[imperial]{Rakesh Heer}
\author[newcastle]{Kabita Adhikari\corref{cor1}}
\cortext[cor1]{Corresponding author}

\affiliation[newcastle]{organization={School of Engineering, Newcastle University}, city={Newcastle upon Tyne}, country={UK}}
\affiliation[southampton]{organization={Electronics and Computer Science, University of Southampton}, city={Southampton}, country={UK}}
\affiliation[urology]{organization={Department of Urology, Freeman Hospital}, city={Newcastle upon Tyne}, country={UK}}
\affiliation[imperial]{organization={Division of Surgery, Imperial College London}, city={London}, country={UK}}

\begin{abstract}

\textbf{Background and Objective:}
Clinicians expect recurrence risk to climb with cancer severity. In a UK multicentre trial, an unconstrained XGBoost model learnt that higher tumour stage and carcinoma in situ predicted lower recurrence risk, and discrimination, calibration, and SHAP were all blind to it. We developed a counterfactual testing framework to detect this inversion and a monotonic-constraint framework to remove it without hurting performance.

\textbf{Methods:}
BOXIT enrolled 472 patients with protocol-mandated cystoscopy across 51 UK sites (2007--2012); 435 had at least two years' follow-up (153 recurrences, 35.2\%). We developed a counterfactual direction test and a monotonic-constraint correction, with constraint directions drawn from the EORTC and EAU risk systems, and evaluated both against unconstrained XGBoost and logistic regression on 18 predictors (seven directed) over 50 cross-validation folds. The test worsened each patient on one directed feature at a time to check whether risk fell; SHAP direction and calibration were also assessed.

\textbf{Key Findings and Limitations:}
Tumour stage and carcinoma in situ were associated with lower recurrence, opposite to medical intuition; the unconstrained model reversed carcinoma in situ counterfactuals in 90.2\% of cases and stage in 74.3\%. Discrimination ($\Delta$AUC 0.005, $p$=0.47), calibration, and SHAP magnitude were all blind to the inversion. Monotonic constraints eliminated every violation at no cost to discrimination (0.723 vs 0.718) and outperformed EORTC ($p$=8.9$\times$10$^{-16}$). Limitations: single trial, internal-external validation only.

\textbf{Conclusions and Clinical Implications:}
A model that had learned this inversion passed every conventional check. A counterfactual direction test, run as a single refit with pre-specified monotonic constraints, catches this failure at no cost to performance and should be routine before clinical deployment.

\end{abstract}


\end{frontmatter}

\vspace{0.5em}
\noindent\textbf{Take Home Message:} A bladder cancer recurrence model learned that worse disease means lower risk, undetected by AUC, calibration, or SHAP. A counterfactual test of feature direction caught it at no cost to accuracy.

\vspace{0.5em}
\noindent\textbf{Patient Summary:} A computer model predicting bladder cancer return had secretly learned that worse cancer meant lower risk, invisible to the checks doctors rely on. We developed a test that catches this mistake and a model that fixes it at no cost to accuracy.

\section{Introduction}

For most patients with bladder cancer, recurrence is simply a matter of when. Non-muscle-invasive bladder cancer returns in up to 78\% of patients within five years \cite{aldousariUpdateManagementNonmuscle2010}, and each recurrence triggers another cystoscopy, another resection, and often another treatment episode, with a cumulative cost of \pounds8{,}735 per patient over three years \cite{coxEffectsBladderCancer2020}. Yet the risk tables most clinicians use to guide surveillance were built in a pre-BCG-maintenance cohort, and in contemporary practice their concordance can fall to 0.554, barely better than chance \cite{xylinasAccuracyEORTCRisk2013}; they also overestimate recurrence in BCG-treated patients \cite{fernandezgomezEORTCTablesOverestimate2011}.

New models are regularly proposed as replacements for the risk tables, reported with AUCs of 0.62 to 0.82 \cite{tangDevelopmentExternalValidation2025,zhangPredictiveValueTotal2024a,schwarzRelevantFeaturesRecurrence2024}. They are judged by the same trio: discrimination, calibration, and a SHAP plot meant to show the model is not a black box \cite{lundbergUnifiedApproachInterpreting2017}. TRIPOD+AI asks for exactly these elements, along with transparent specification and fairness assessment \cite{collinsTRIPOD+AIStatementUpdated2024}. A model can perform well on all three and still learn the wrong direction of effect. For example in BOXIT, the UK NMIBC trial data used in this study, patients with T1 disease or carcinoma in situ, both more advanced than Ta, were assigned lower recurrence probabilities than Ta patients. No standard check catches this, because none asks whether predicted risk rises or falls with each feature. They measure only how much each feature matters and how accurate the model is overall. 

The same pattern recurs in PHOTO, another UK NMIBC trial, where T1 patients recurred less often than Ta patients (Supplementary Section S13), hence the model learnt exactly that. In prostate cancer, a model learned that worse stage predicted better survival \cite{aligningPredictionClinicalLearning2025}. These reversals are consistent with the treatment paradox: higher-risk patients receive more intensive treatment, so the risk factor appears protective \cite{cheongSeeTreatmentParadox2016}. Beyond urology, pneumonia and cardiovascular models learned that asthma and treated hypertension were protective \cite{caruanaIntelligibleModelsHealthCare2015,vanGelovenPredictionMeetsCausal2020}, and glass-box models on four further datasets learned that aggressive treatment lowers apparent risk \cite{lengerichDeathRoundNumbers2022}. Such reversals are found only by chance, and no established method exists to detect them systematically.

In this paper, we developed a two-part framework. First, a counterfactual test examines the model's understanding of each feature, checking whether predicted risk moves in the clinically expected direction. Second, monotonic constraints enforce that direction wherever the test finds a violation. To our knowledge, this framework has not previously been applied in urological cancers. We develop and compare models in BOXIT, validate them geographically across its 51 UK recruiting sites, and examine whether reversals are explained by treatment intensity. Our aim is to give clinicians a simple test for any recurrence model, whether predicted risk rises as disease gets worse. We show how to make a Machine learning model learn the correct direction without losing accuracy.

\section{Methods}

\subsection{Study Design and Data Source}

\begin{figure}[htbp]
\centering
\includegraphics[width=\textwidth]{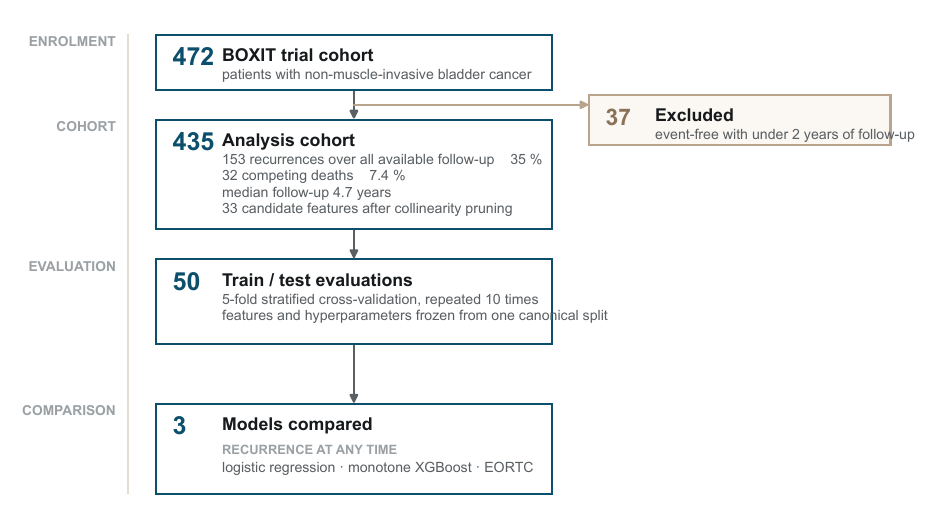}
\caption{Study cohort and analysis design. Patients are labelled by whether recurrence was ever observed; those event-free with under two years of follow-up are excluded, since they have not completed the EAU-guideline period of closest surveillance for high-risk disease. The 35.2\% recurrence rate in the analysis cohort is higher than the trial's overall rate because this exclusion removes only event-free patients, never patients who recurred.} \label{fig:cohort}
\end{figure}

Data came from the BOXIT trial (ISRCTN 84311762), a prospective UK multicentre randomised trial of intermediate- and high-risk NMIBC, which enrolled 472 patients across 51 sites between November 2007 and July 2012 with protocol-mandated cystoscopic surveillance; recurrence was assessed on a fixed schedule by local pathologists blinded to any later model. \Cref{fig:cohort} summarises enrolment, the analysis cohort, and the evaluation design.

\subsection{Participants and Outcome}

The primary outcome was histologically confirmed recurrence at any point during follow-up. Event-free patients with under two years' follow-up were excluded: EAU-guideline surveillance for high-risk NMIBC is three-monthly cystoscopy and cytology for the first two years \cite{eauFollowUpNMIBC}, so patients below this floor cannot be reliably labelled event-free. This removed 37 of 472 enrolled patients, leaving an analysis cohort of 435 (153 recurrences, 35.2\%; 32 competing deaths, 7.4\%; median follow-up 4.7 years).
\subsection{Clinical Direction Priors}
\label{sec:priors}

Every constraint direction is drawn from published, guideline-endorsed risk systems: the EORTC risk tables \cite{sylvester2006} and EAU risk stratification \cite{sylvester2021,PredictingDiseaseRecurrence}. Seven features were constrained (tumour grade, stage, number, diameter, recurrence rate, carcinoma in situ, and age) so that higher values cannot lower predicted risk. Recurrence rate is prior recurrences in the two years before registration, per year, excluding the registration tumour itself; the past-year count stays in the model, unconstrained. EORTC was scored from the trial's coded items \cite{sylvester2006}.

\subsection{Feature Selection and Modelling}

Boruta feature selection, with the directed features always retained, gave 18 predictors (\Cref{sec:priors}) after excluding composites that re-encode stage or carcinoma in situ, which could bypass the constraints, and induction dose counts and durations, which shorten on early recurrence; any induction treatment was kept. An XGBoost classifier \cite{chenXGBoostScalableTree2016,guptaMonotonicCalibratedInterpolated2016} was fitted with these monotonic constraints; an identical classifier was refitted with the constraints removed and nothing else altered, and an L2-regularised logistic regression on the same features served as an interpretable comparator. Hyperparameters were tuned once on a fixed 80:20 split, then frozen; all reported performance comes from repeated stratified five-fold cross-validation (10 repeats, 50 evaluations).

\subsection{The Counterfactual Direction Test}

For every patient and every clinically-directed feature, we moved that feature to the next-worse value actually observed in the cohort (for example, tumour stage Ta to T1), with everything else held fixed, and checked whether the model's predicted recurrence probability fell: a violation, since worse disease should not predict lower risk. This tests the model's output directly, rather than inferring direction from a feature-importance summary. We ran the test in all 50 cross-validation folds on the constrained model, the unconstrained model, and logistic regression, separating a violation found once from one that recurs on every partition. We separately report SHAP attribution direction (Spearman correlation between feature value and attribution) and partial dependence, since Shapley-value attributions are known to encode which features drove a prediction without reliably encoding which way they pushed it \cite{kumarProblemsShapleyvaluebasedExplanations2020}. We also compared SHAP attributions for individual held-out patients between the unconstrained and constrained fits.

\subsection{Statistical Analysis}

Discrimination is reported as mean (SD) AUC across the 50 folds, with paired within-fold comparisons (Wilcoxon signed-rank) against EORTC and logistic regression. Sample-size adequacy was assessed against Riley et al.'s criteria for a binary prediction model \cite{rileyMinimumSampleSize2019}. Weak calibration (slope, intercept, observed:expected ratio) was assessed on pooled out-of-fold predictions before and after cross-fitted Platt recalibration \cite{vancalsterCalibrationAchillesHeel2019}. A parsimony sweep refitted the model at decreasing feature-set sizes, ranked by pooled SHAP importance, on the same 50 folds.

\section{Results}

\subsection{The cohort's own biology runs backwards}

Independent of any model, observed recurrence in this cohort falls as tumour stage worsens: 45.1\% at Ta ($n$=204), 28.4\% at T1 ($n$=183), 18.8\% at Tis ($n$=48), the opposite of guideline teaching, and not an artefact of carcinoma in situ: among patients without it, Ta recurred in 47.4\% and T1 in 28.3\% ($p$<0.001). Number of tumours and recurrence rate are the directed features whose marginals agree with clinical teaching, making them the decisive test below.

\subsection{The unconstrained model learned the inversion}

Across the 50 folds, worsening carcinoma in situ lowered the unconstrained model's predicted risk in 90.2\% of counterfactuals (worst case $-$20.2 points; \Cref{tab:audit}) and raising tumour stage one level did so in 74.3\% (worst $-$21.8 points). Averaged across all seven directed features, the unconstrained model reversed 47.5\% of counterfactuals (SD 2.7, range 40.1--53.6\%) against 0.0\% for the constrained model by construction ($p$<0.001). Logistic regression was no safer: its carcinoma in situ coefficient was negative in 50 of 50 folds, its grade coefficient in 47 and its stage coefficient in 41, a mean violation rate of 41.1\%, below XGBoost's 47.5\% but bimodal rather than partial (carcinoma in situ 100\% in every fold). An unconstrained random forest reversed at a similar rate (Supplementary Table S1).

\begin{table}[htbp]
\centering
\caption{Counterfactual direction test of the seven clinically-directed features, mean over the 50 cross-validation folds, each audited on its training partition. Each patient's feature is worsened by one observed level with everything else held fixed; the column reports the share of such counterfactuals for which predicted recurrence probability fell, and the largest fall is the worst single counterfactual across all folds. Under monotone constraints this share is zero for every feature by construction, so the constrained column is omitted. SHAP attributions carry no such guarantee, so both columns are measured.}
\label{tab:audit}
\small
\setlength{\tabcolsep}{4.5pt}
\begin{tabular}{lrrrrrr}
\toprule
& \multicolumn{2}{c}{Counterfactual test} & \multicolumn{2}{c}{SHAP violations, \%} & \multicolumn{2}{c}{Spearman $\rho$} \\
Feature & Risk fell, \% & Largest fall, pp & Unconstr. & Constr. & Unconstr. & Constr. \\
\midrule
Carcinoma in situ         & 90.2 & $-$20.2 & 99.7 & 0.0 & $-$0.66 & +0.67 \\
Tumour stage              & 74.3 & $-$21.8 & 94.7 & 2.1 & $-$0.85 & +0.88 \\
Tumour grade              & 74.3 & $-$17.5 & 91.3 & 0.4 & $-$0.78 & +0.91 \\
Age                       & 40.0 & $-$20.4 & 38.9 & 10.8 & +0.34 & +0.91 \\
Tumour diameter           & 26.6 & $-$6.6 & 10.8 & 5.5 & +0.87 & +0.92 \\
Number of tumours         & 19.7 & $-$11.9 & 8.4 & 4.5 & +0.84 & +0.87 \\
Recurrence rate           & 7.6 & $-$10.8 & 1.6 & 0.6 & +0.78 & +0.79 \\
\bottomrule
\end{tabular}
\end{table}

The decisive result is on number of tumours: the unconstrained model reversed it anyway, in 19.7\% of counterfactuals (worst $-$11.9 points), so this cannot be excused as a correctly-learned inverted cohort; recurrence rate, whose marginals also agree, was reversed less often (7.6\%). Supplementary Figure S7 shows predicted risk by stage against observed recurrence and the per-fold test spread across all three model classes.

\subsection{Nothing in standard validation caught it}

Discrimination barely moved between the two fits: mean AUC 0.718 (unconstrained) versus 0.723 (constrained), a mean paired difference of 0.005 ($p$=0.47). Calibration gave no warning either: the uncalibrated slope (1.13) and observed:expected ratio (0.80) describe a merely over-predicting model, and after cross-fitted Platt recalibration the slope moved to 1.00 with AUC unchanged (0.718) while the inversion persisted underneath a perfectly calibrated surface (Supplementary Table S2), a gap others have warned calibration alone cannot close \cite{vancalsterCalibrationAchillesHeel2019}.

\begin{figure}[htbp]
\centering
\includegraphics[width=0.92\textwidth,height=0.78\textheight,keepaspectratio]{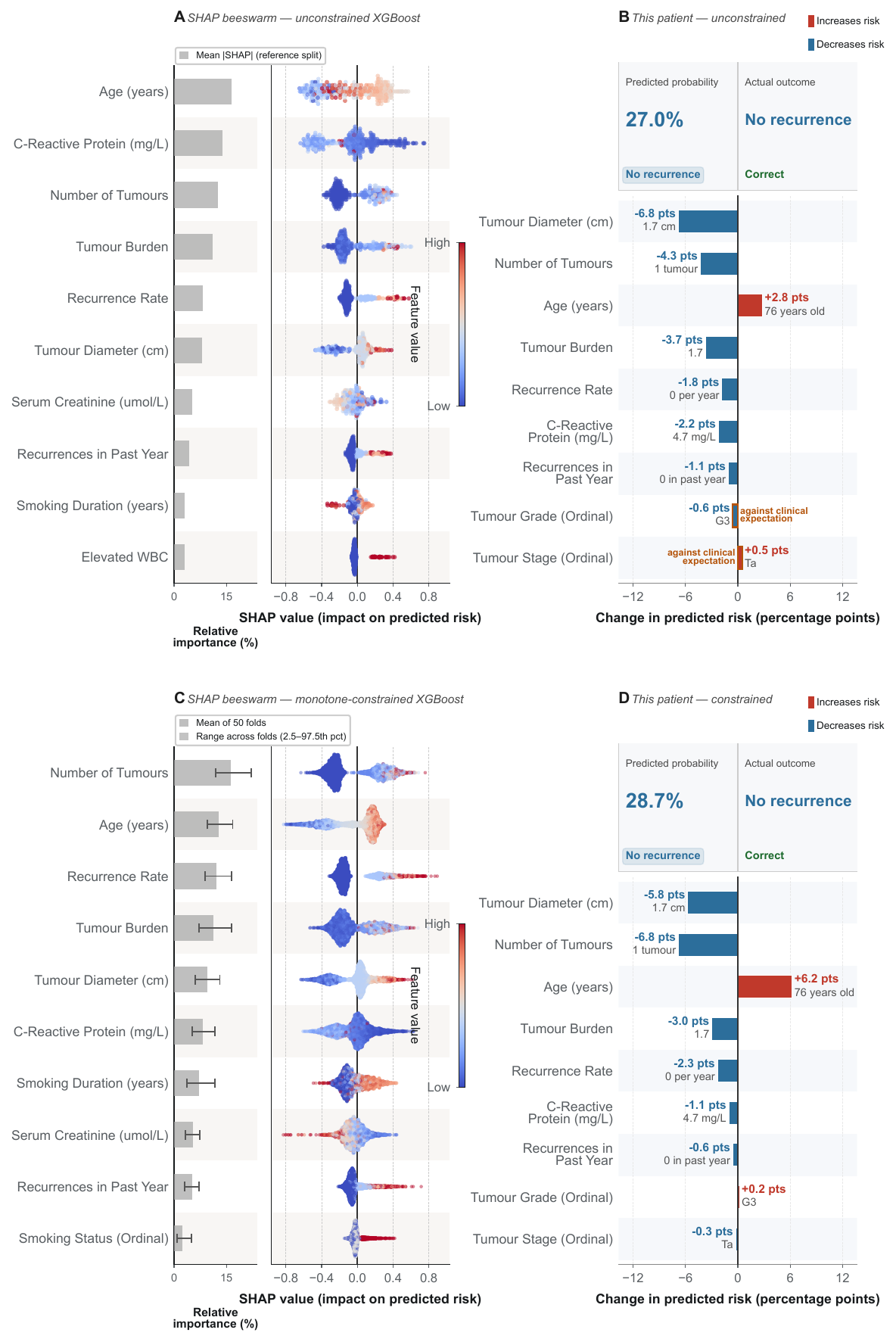}
\caption{Post-hoc SHAP explanation of the unconstrained (A, B) and monotone-constrained (C, D) XGBoost fits. (A, C) Beeswarm of the ten highest-ranked predictors, coloured by feature value; bars show mean $|$SHAP$|$ on the single reference split in A and the mean across the 50 folds with its 2.5--97.5th percentile range in C, where the dots pool all 50 folds. (B, D) Signed contributions, in percentage points of predicted risk, for one held-out patient (stage Ta, grade G3, one tumour of 1.7 cm, age 76, no recurrence) chosen by pre-specified rules from 87 test patients (Supplementary Section S12). The unconstrained fit assigns grade $-$0.6 and stage $+$0.5 points (highlighted, against clinical expectation) and predicts 27.0\%; the constrained fit assigns $+$0.2 and $-$0.3 points and predicts 28.7\%. A post-hoc attribution can only describe what a fitted model does, not whether the direction of its effect is clinically defensible.}
\label{fig:shap-lime}
\end{figure}

Post-hoc explanation gave no warning either. \Cref{fig:shap-lime} sets SHAP for the unconstrained fit (A, B) beside the constrained fit (C, D), each as a global beeswarm and as a decomposition for one held-out patient. Both global plots show a plausible, legible top ten. In neither fit do tumour stage or carcinoma in situ reach the top ten, and nothing on the importance bars says their attributions push risk down.

Restricted to the seven directed features (Table 1), SHAP flags stage regardless of direction (94.7\% unconstrained, 2.1\% constrained; Spearman $\rho$ $-$0.85 to +0.88), the same signal at opposite signs, indistinguishable on an importance plot that reports only magnitude, the failure mode Shapley-based explanations are known to carry \cite{kumarProblemsShapleyvaluebasedExplanations2020}. Supplementary Figure S8 confirms neither SHAP magnitude nor discrimination separates the fits enough to catch the reversal.

\subsection{Constraints fix it, and cost nothing}

\begin{figure}[htbp]
\centering
\includegraphics[width=3.6in]{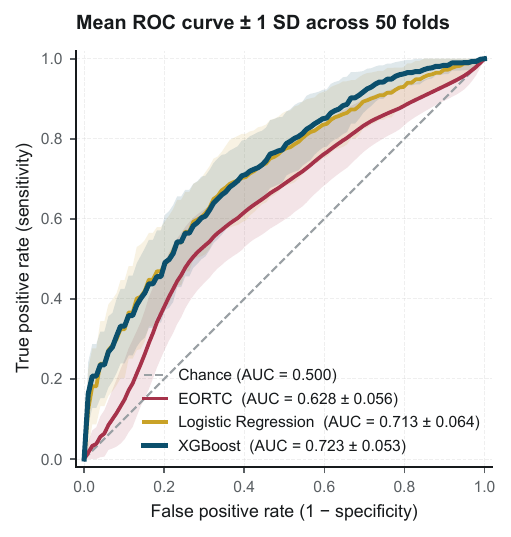}
\caption{Mean ROC curve $\pm$ 1 SD across the 50 cross-validation folds for the constrained model, logistic regression, and EORTC.}
\label{fig:roc}
\end{figure}

Applying the seven monotonic constraints reduced every detected violation to zero, by construction, at no cost to discrimination. The constrained model outperformed EORTC (mean AUC 0.723 versus 0.628; wins in 50 of 50 folds; Wilcoxon $p$=8.9$\times$10$^{-16}$) and matched logistic regression (0.723 versus 0.713; $p$=0.14), as \Cref{fig:roc} shows.

\begin{figure}[htbp]
\centering
\includegraphics[width=\textwidth]{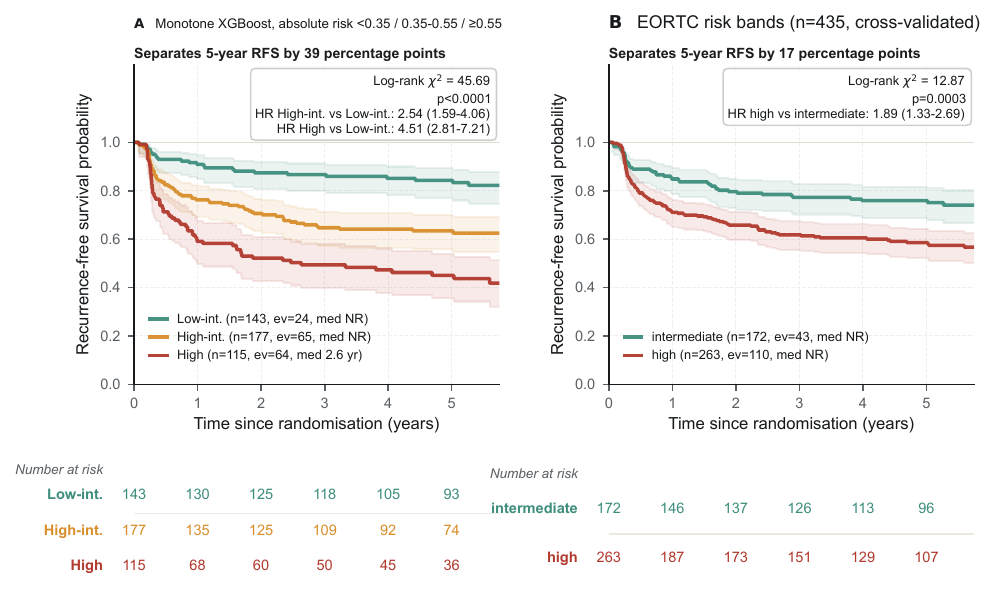}
\caption{Recurrence-free survival by risk tier, out-of-fold across all 50 cross-validation folds (n=435). (A) Constrained model risk groups by absolute predicted risk, <0.35 / 0.35--0.55 / $\geq$0.55 (Low-intermediate / High-intermediate / High; no BOXIT patient is truly low-risk under EAU criteria). (B) EORTC published bands on the same patients; the low-risk band is empty in this cohort. Log-rank statistics, hazard ratios, and number at risk are on the panel.}
\label{fig:km}
\end{figure}

Across the same folds, grouping by absolute predicted risk (<0.35, 0.35--0.55, $\geq$0.55; $n$=143/177/115), the constrained model separates recurrence-free survival more than EORTC's published bands (39.3 versus 17.3 points; $\chi^2$=45.69 versus 12.87, $p$<0.0001 and $p$=0.0003; HR High vs Low-intermediate 4.51, 95\% CI 2.81--7.21; \Cref{fig:km}). At a threshold of 0.30 it gives 31\% higher net benefit than EORTC (0.134 versus 0.102) and spares 7 more cystoscopies per 100 patients (Supplementary Figure S5), and separates recurrence risk within a single EORTC band (Supplementary Figure S6).

Tumour stage falls to a median importance rank of 17 of 18 (median mean $|$SHAP$|$ 0.010), tumour grade to 15 of 18 (0.018) and carcinoma in situ to 18 of 18 (0.003) across the 50 folds, none entering the top ten in any fold, while the model rebuilds equivalent discrimination from the eleven unconstrained predictors (Supplementary Figure S1). The constraint does not reorient these three features; it silences them. That is correct behaviour, since a protective stage association produced by treatment assignment will not survive contact with a clinic where treatment is assigned differently; a model carrying no stage effect is safer to deploy than one carrying an inverted one, at no cost in AUC. The effect is also selective: it binds hardest where cohort and clinical prior conflict, leaving alone recurrence rate, age, number of tumours and tumour diameter, which take four of the five top places.

\subsection{The deployable model can be smaller}

By Riley et al.'s criteria \cite{rileyMinimumSampleSize2019}, an 18-predictor binary model needs 1{,}454 patients to bound shrinkage at 10\% (events per parameter 8.5, Cox-Snell $R^2$=0.105); this cohort has 435, supporting roughly 5.4 predictors, a shortfall the events-per-variable rule would not flag \cite{peduzziSimulationStudyNumber1996}. A parsimony sweep finds no rung from 18 down to five predictors significantly worse than the best (five, 0.727, versus 0.723 for all 18) except eight (0.713, $p$=0.032), with a sharp fall at four (0.705) and three (0.682) (Supplementary Table S3).

\subsection{The inversion is not explained by the follow-up exclusion}

The two-year follow-up floor is a natural first suspect, since it selectively removes event-free patients and can bias the remaining cohort toward lower-risk stage patterns. Fitting the stage-recurrence association three ways (full 472-patient cohort, 435-patient analysis cohort, and the audit's logistic model; Supplementary Figure S4), all three give a hazard or odds ratio below 1 for higher stage (full cohort HR 0.59, 95\% CI 0.45--0.77; analysis cohort HR 0.58, 95\% CI 0.45--0.76; OR 0.51, 95\% CI 0.37--0.70), none crossing 1: the inversion predates any follow-up-based exclusion, so the exclusion does not explain it.

\subsection{The constraint preserves geographic transportability}

Leave-one-site-out validation shows the constrained model transfers with a wide spread (pooled out-of-site AUC 0.721 versus 0.627 for EORTC, median per-group AUC 0.733, range 0.529--0.825, exceeding EORTC in 8 of 9 site-groups), and the same constrained-versus-unconstrained comparison at the site-group level shows no significant difference (pooled AUC 0.721 versus 0.721; per-group Wilcoxon $p$>0.99): directional coherence comes at no measurable cost to transfer.

\section{Discussion}

\subsection{Why the inversion shows in the unconstrained model}

Selection bias \cite{kristmanLossFollowUpCohort2004,levesqueImmortalTimeBias2010} is ruled out: the inversion is present in the full cohort before any exclusion.

The likelier driver is treatment by indication: treatment intensity was not assigned at random with respect to severity, and patients with more advanced stage or carcinoma in situ plausibly received more aggressive therapy, suppressing the recurrence their stage would otherwise predict. We cannot fully separate treatment intensity from severity in this dataset, but the inversion surviving every cohort policy tested makes exclusion the weaker candidate. Either way, an unconstrained learner fits whatever association is present, including one produced by clinical decision-making rather than biology \cite{zechVariableGeneralizationPerformance2018}. Consistent with this, under maintenance BCG prior recurrence rate and tumour number predict recurrence while stage and grade predict progression and death \cite{cambierEORTCNomogramsRisk2016}. Treatment that begins after baseline distorts prediction models in a related way \cite{sperrinUsingMarginalStructural2018}, and induction dose counts, which shorten on early recurrence, were excluded for that reason. As raised in the Introduction, the same mechanism is documented in PHOTO and beyond bladder cancer (Supplementary Section S13).

\subsection{It happened in a clean trial}

BOXIT is a prospective trial with protocol-mandated cystoscopy, about as favourable a setting as this data gets, yet the inversion still appeared and survived discrimination, calibration, and SHAP undetected. Registry and routine hospital records, with less complete follow-up, are at greater risk of the same failure, not less.

\subsection{Clinicians distrusting these models are not confused}

When a clinician rejects a model's output because it disagrees with what they know about disease progression, that reaction is often read as resistance to change; here it would have been a correct catch. Explaining a model's output verifies that a feature matters, not that it matters in the right direction, a distinction close to Rudin's argument that explaining a black box is no substitute for constraining one \cite{rudinStopExplainingBlack2019,liptonMythosModelInterpretability2018}. The clearest precedent is Caruana and colleagues' pneumonia model, which learned that asthma lowered readmission risk because asthmatic patients were triaged more aggressively, caught only because the model's structure could be read directly off the page \cite{caruanaIntelligibleModelsHealthCare2015}; gradient-boosted tabular models are not read that way. Our recommendation is narrow and cheap: add a directional counterfactual audit alongside calibration, one refit against pre-specified clinical priors that takes an afternoon.

\subsection{Limitations}

This is one trial, 435 patients, enrolled 2007--2012, evaluated on internal cross-validation; independent external validation is required before clinical use. EORTC was scored from the trial's coded items, not the published tables, so comparisons with it are optimistic. Neither the inverted coefficients nor the constraint are causal claims \cite{westreichTable2Fallacy2013}: the constraint limits what the model may say and does not estimate a treatment effect.

The treatment-by-indication account above is consistent with every cohort policy tested but not directly demonstrated, since this trial has no treatment-intensity variable independent enough of disease severity to test it head-on (Supplementary Figure S4). Two further checks were inconclusive: inversion severity was unrelated to site-level dropout, and while stage and grade reversal rose with the label horizon, the other features did not (Supplementary Figures S2 and S3).

\section{Conclusion}

Unconstrained models can learn relationships that contradict established medical knowledge. Trained on a clean prospective trial, an XGBoost model learned that higher tumour stage and carcinoma in situ predict lower recurrence risk, an inversion of basic pathology. This passed discrimination, calibration, and SHAP undetected. Monotonic constraints, applied to the same model, removed it at no cost to performance \cite{guptaMonotonicCalibratedInterpolated2016}. We believe the counterfactual direction test that caught it is the more durable contribution: a single refit that belongs in routine validation before any clinical model is deployed.

\bibliographystyle{elsarticle-num}
\bibliography{XGBoost_references}

\end{document}